\documentclass[11pt]{article}
\usepackage{nodalida2025}
\usepackage{times}
\usepackage{url}
\usepackage{latexsym}

\usepackage[T1]{fontenc}
\usepackage{graphicx}
\usepackage{hyperref}
\usepackage{xcolor}
\hypersetup{
    colorlinks,
    linkcolor={black!50!black},
    citecolor={black!50!black},
    urlcolor={black!80!black}
}
\usepackage{enumitem}
\usepackage{amsmath}
\usepackage[singlelinecheck=false,justification=raggedright]{caption}
\usepackage{subcaption}

\NewDocumentCommand{\rot}{O{35} O{1em} m}{\makebox[#2][l]{\rotatebox{#1}{#3}}}%
\NewDocumentCommand{\rotmore}{O{70} O{1em} m}{\makebox[#2][l]{\rotatebox{#1}{#3}}}%

\aclfinalcopy 

\title{Profiling Bias in LLMs: \\ Stereotype Dimensions in Contextual Word Embeddings}

\author{Carolin M. Schuster, Maria-Alexandra Dinisor, Shashwat Ghatiwala \and Georg Groh \\
         TUM School of Computation, Information and Technology
         \\ Technical University of Munich \\ 
         {\tt \{carolin.schuster, alexandra.dinisor, shashwat.ghatiwala\}@tum.de} \\
         {\tt grohg@in.tum.de}
         }

\date{}

\begin{document}
\maketitle

\global\csname @topnum\endcsname 0
\global\csname @botnum\endcsname 0

\begin{abstract}
Large language models (LLMs) are the foundation of the current successes of artificial intelligence (AI), however, they are unavoidably biased. To effectively communicate the risks and encourage mitigation efforts these models need adequate and intuitive descriptions of their discriminatory properties, appropriate for all audiences of AI. We suggest bias profiles with respect to stereotype dimensions based on dictionaries from social psychology research. Along these dimensions we investigate gender bias in contextual embeddings, across contexts and layers, and generate stereotype profiles for twelve different LLMs, demonstrating their intuition and use case for exposing and visualizing bias.
\end{abstract}

\section{Introduction}

\begin{figure}
   
    \includegraphics[width=0.5\textwidth]{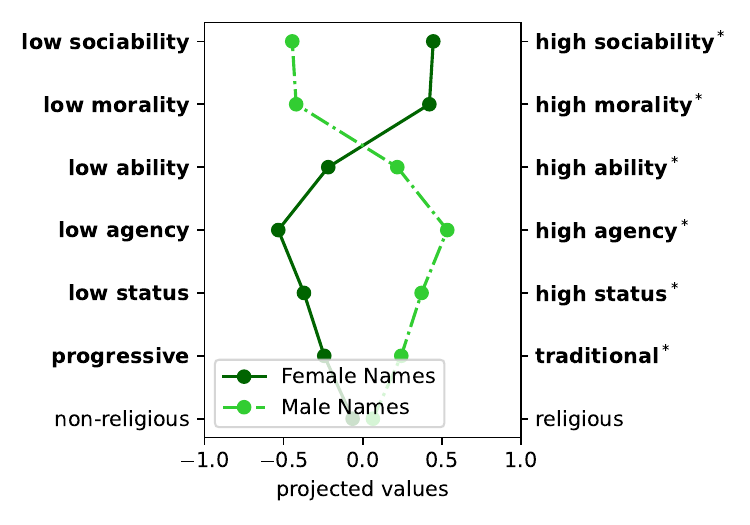}
     \caption{7D stereotype profile for Llama-3-8B, revealing differences in embeddings of 100 female and 100 male-associated names. \textbf{*}Statistically significant differences (p<0.05).}
    \label{fig:cover_plot}
\end{figure}

\begin{table*}[ht]

\begin{tabular}{r| l l l l l l l  }
 dimension  &  direction & n  & terms & $\textrm{n}_{add.}$ & additional terms
 \\

         \hline
 sociability &  high & 43 & nice, friendliness, warmth & 199&  accomodating, witty \\
  &  low & 42 &  unfriendly, unsociability, distant & 162 & acid, withdrawn \\
 \hline
 morality & high & 51 &   humane, morality, benevolent & 205 &  allegiance, true \\
  & low & 69 & untrustworthiness, evil, insincere & 635 & abandon, wrongful  \\
 \hline
 ability & high  & 40  & intelligence, capable, graceful & 302 & accomplished, ace \\
   & low & 39  &   ignorant, stupid, inefficient & 160 & awkward, unadvised\\
 \hline
 agency & high & 42 & motivated, autonomous, resolute & 256 & action, worker\\
  & low & 39 & vulnerable, submission, helpless & 113 & bowing, unsure \\
 \hline
 status & high & 21 &    superior, wealth, important & 187 & advantage, win \\
  & low & 13 &  poor, insignificant, unsuccessful &  117 & bankrupt, welfare \\
 \hline
 politics  & traditional & 12 &   conventional, conservative & 34 & classical, capitalist  \\
  & progressive & 16 &  modern, liberal,  democrat & 45 & contemporary, feminist\\
 \hline 
 religion & religious & 18 &  believer, church, god-fearing & 146 & spirit, testament\\
  & non-religious & 10 & atheist, skeptical, secular & 6 & unholy, impious\\
         
\end{tabular}

\caption{Examples of terms and their directions 
on stereotype dimensions from the theoretically grounded dictionary by \citet{Nicolas2021}. The additional terms were collected from their extended dictionary created by a semi-automated method. High-level stereotype dimensions are constructed as follows: warmth = sociability + morality, competence = ability + agency.} 
    \label{tab:word_counts}

\end{table*}

Amongst many other semantic concepts, large language models (LLMs) pick up stereotypes from the data they are trained on. Unbiased data is hard to come by, especially in the amounts needed for the ever-larger models, which are the foundation of the current successes of AI and the respective hype. Thus bias in these models is basically unavoidable, making it necessary to understand its characteristics and extents to communicate the risks and find ways to mitigate adverse discriminatory effects on affected populations. 

Past research on bias has often involved word embedding association tests \citep{Caliskan2017}, inspired by the implicit association tests (IAT) \citep{Greenwald1998} of social psychology. By another inspiration from the social sciences, a newer direction of natural language processing (NLP) research transforms opaque embeddings into a space of meaningful dimensions \citep{Mathew2020, Kwak2021, Senel2022, engler-etal-2022-sensepolar}, enabling new ways to study concepts. Similar to semantic differentials \citep{Osgood1957}, this methodology relies on antonyms (e.g. fast vs. slow) or opposing concepts described by lexicons.

In this work we study bias in LLMs by transforming their embeddings based on the stereotype content model (SCM) by \citet{fiske2002model}, enabling the study along theoretically and empirically grounded stereotype dimensions. The SCM entails two primary dimensions originating from interactions, where people seek to understand the other party's intent (dimension of warmth) and their capabilities (dimension of competence). Stereotypically, women are thereby associated with higher warmth, and men with higher competence. Furthermore we employ the extended model \citep{Abele2016, Ellemers2017, Goodwin2015, Koch2016}, allowing us to provide detailed 7D bias profiles as shown in \autoref{fig:cover_plot}.

\paragraph{Contributions.} In summary, we \textbf{(i)} show how the stereotype content model can be employed to expose and visualize bias in contextual embeddings\footnote{Code available at \url{https://github.com/carolinmschuster/profiling-bias-in-llms}}, \textbf{(ii)} generate bias profiles for twelve LLMs for gender-associated names and gendered terms, displaying overall stereotypical associations of warmth and competence, \textbf{(iii)} provide insights on stereotype dimensions and gender bias across context examples and network layers.

\section{Related Work}

Inspired by the human implicit association test \citep{Greenwald1998}, \citet{Caliskan2017} developed the first Word Embedding Association Test (WEAT) to assess the association between two target concepts (e.g., scientist vs. librarian) and two attributes (e.g., male vs. female) in static word embeddings by cosine similarity and a permutation test. Later \citet{Tan2019} built a first approach to measure bias for LLMs using the contextual embeddings of the words within examples. The Contextual Embedding Association Test (CEAT) \citep{Guo2021a} employs a random effects model to quantify bias with sampled contexts from a corpus.

A newer approach to interpreting the high-dimensional embedding spaces is again inspired by a concept from the social sciences; semantic differentials \citep{Osgood1957}. \citet{Mathew2020} introduced POLAR, a transformation of static word embeddings to a new polar, interpretable space. The polar opposites are antonyms such as hot-cold or soft-hard, and their word vectors are employed to define the new dimensions, which were shown to align with human judgment in an evaluation study. Similar frameworks are SemAxis \citep{An2018}, FrameAxis \citep{Kwak2021} and BiImp \citep{Senel2022}.

More recently, the SensePolar framework was introduced by \citet{engler-etal-2022-sensepolar}, extending the POLAR approach to contextual word embeddings. The poles are hereby defined not by the word alone but by their embedding within sense-specific example sentences from a dictionary. 
The authors showed that these more interpretable embeddings can achieve similar performance to regular ones on natural language understanding (NLU) tasks, and furthermore confirmed the approach by a human evaluation study.

Most similar to our work \citet{fraser-etal-2021-understanding} analyzed stereotype dimensions in static embeddings, combining the POLAR framework by \citet{Mathew2020} with the warmth and competence dimensions of the stereotype content model \citep{fiske2002model}. They demonstrated that static word embeddings can recreate the stereotype dimensions from literature by predicting the cold-warm and competent-incompetent associations for additional known words, and by further comparing the results to psychological surveys.

For contextual embeddings \citet{ungless-etal-2022-robust} measured bias with CEAT \citep{Guo2021a} based on the warmth and competence dimensions and in a generation-based approach
\citet{jeoung-etal-2023-stereomap} elicited evaluation of different social groups on these dimensions, with multiple prompting strategies. The stereotype content model has also been used for de-biasing methods \citep{ungless-etal-2022-robust, omrani-etal-2023-social}. The researchers suggest that this theory-driven approach has an advantage because it is social-group-agnostic and thus does not require iteration over discriminated groups or previous knowledge of specific bias.

In another projection approach, \citet{omrani2023evaluating} used a maximum margin support vector classifier to learn the valence subspace (pleasantness vs. unpleasantness) and projected the word `person' to this dimension, placing different words in its context. The bias between words is measured by their effect on the contextualized representation of `person'.

This work furthermore relates with a broader range of studies trying to understand the contents of contextual representations, most notably by knowledge probing (e.g. \citet{tenneyyou, schuster-hegelich-2022-berts}). See \citet{cao2024life} for a recent survey.

\section{Experimental Setup}

\subsection{Stereotype Dimensions \& Dictionaries}

Our analysis of stereotype dimensions and bias in LLMs is grounded in the stereotype content model by \citet{fiske2002model}, who showed that there are two major dimensions of warmth and competence and that many stereotypes are mixed along these two. 
Following \citet{Nicolas2021} we also study the more fine-grained dimensions of sociability and morality for warmth \citep{Abele2016} and ability and agency for competence \citep{Ellemers2017, Goodwin2015}. Expanding the set of concepts by the Agency-Beliefs-Communion model  \citep{Koch2016}, we further include the dimensions of status, politics, and religion. This allows us to also provide a more detailed and extended stereotype profile for the LLMs.

Akin to previous work \citep{fraser-etal-2021-understanding, omrani-etal-2023-social} we use the dictionaries in the supplementary data from \citet{Nicolas2021}\footnote{https://osf.io/yx45f/}, which were validated by human evaluation studies, for the construction of the stereotype space. 

The `seed dictionary' is theory-driven, using terms from literature, while the `full dictionary' contains additional terms collected by a semi-automated method, identifying synonyms using the English lexical database WordNet \citep{Miller1990}\footnote{https://wordnet.princeton.edu/}.
Both dictionaries distinguish seven stereotype dimensions, as shown in \autoref{tab:word_counts} with examples of terms per dimension and direction. 
While most dimensions are coded low--high, the politics dimension is coded progressive--traditional and the dimension of religion is coded non-religious--religious.

\begin{figure}[ht!]
    \centering
    \includegraphics[width=0.5\textwidth]{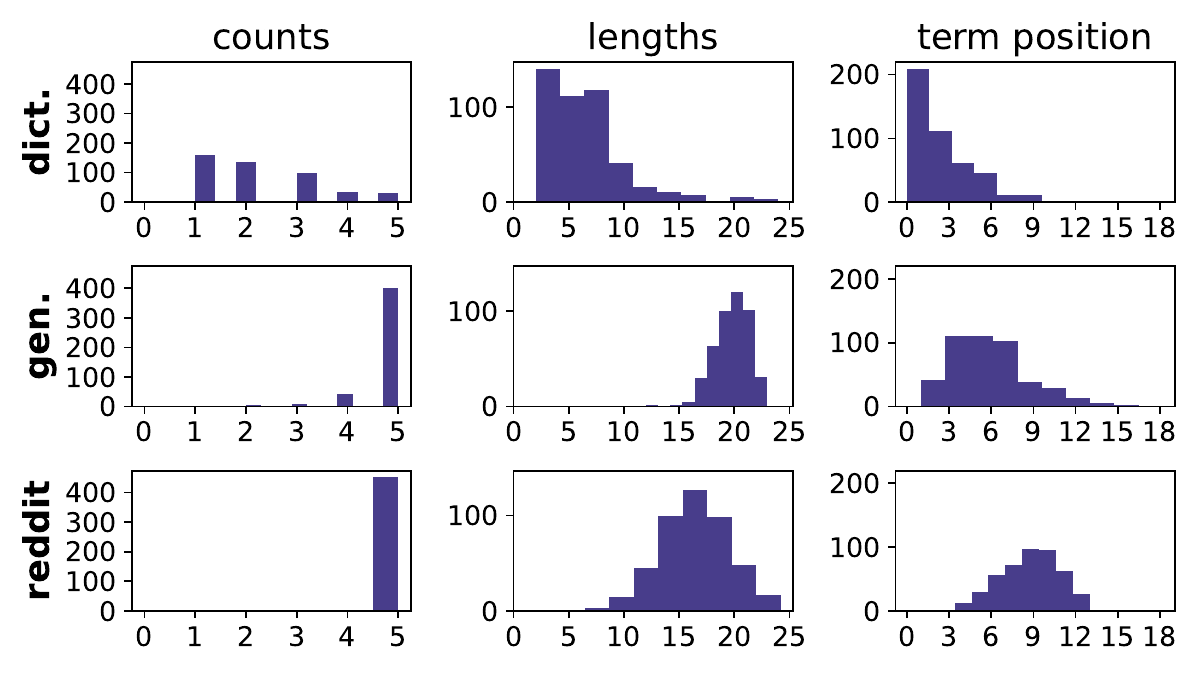}
    \caption{Properties of context examples: Histograms of example counts, numbers of words and positions of dictionary terms within the examples.}
    \label{fig:example__for_seed_terms}
\end{figure}

\subsection{Context Examples} \label{context_examples}

As we are working with contextual embeddings the context of the terms becomes a crucial design choice for the study of stereotype dimensions and bias (see also \citet{engler-etal-2022-sensepolar}). 

\textbf{Generated Examples:}
For our main experiments, we generate gender non-specific contexts with Llama-3-8B-instruct \cite{llama3modelcard, dubey2024llama} by the instruction to avoid names and gender-specific pronouns. Thus, no additional gender bias is introduced. 
As synset information is available for the terms in the seed dictionary, the prompts for these terms additionally include the term definition from WordNet \citep{Miller1990}, allowing a more precise generation for the specific word meaning. 

\textbf{Dictionary Examples:} As in the original SensePolar paper \citep{engler-etal-2022-sensepolar}, we also retrieve context examples from WordNet \citep{Miller1990}, and only for the seed stereotype dictionary do we manually add examples from 
other dictionaries where WordNet does not provide any.  

\textbf{Reddit Examples:}
We include one setup with natural data, where we sample term examples from a Reddit Corpus\footnote{https://www.kaggle.com/datasets/kaggle/reddit-comments-may-2015}, similar as done in the Contextual Word Embedding Association Test \citep{Guo2021a}. 

\textbf{No Context:} In this setting only the terms are passed through the models, preventing contextualization beyond term subwords. 

\vspace{0.1cm}

Example properties are shown in \autoref{fig:example__for_seed_terms}. 
We set the number of examples to five for comparison and to limit computational time, but there often are fewer available for the dictionary examples. Dictionary examples are also the shortest, as they are often short phrases, e.g. ``friendly advice''. The generated examples are the longest with an average of 20 words. Reddit examples are truncated on both sides to include context around the term, which may explain later term positions.

\begin{table*}[ht]

\resizebox{\textwidth}{!}{

\begin{tabular}{ l  l l l l l l l l l}
 & \rot{Warmth} & \rot{Competence} & \hspace{0.1cm}\rot{Sociability} & \rot{Morality} & \rot{Ability} & \rot{Agency} & \rot{Status} & \rot{Politics} & \rot{Religion}  \\

\hline
Llama-3-8B \citep{llama3modelcard}  & \vline \hspace{0.1cm} 0.79 & 0.81 &  \vline \hspace{0.1cm} 0.66 & 0.87 & 0.72 & 0.77 & 0.8 & 0.73 & 0.74 \\
Llama-3-8B-Instruct \citep{llama3modelcard}& \vline \hspace{0.1cm} 0.79 & 0.82 & \vline \hspace{0.1cm} 0.65 & \textbf{0.88} & 0.75 & 0.77 & \textbf{0.83} & 0.83 & 0.81 \\
Llama-3.2-3B \citep{llama3.2modelcard}& \vline \hspace{0.1cm} 0.81 & 0.8 & \vline \hspace{0.1cm} 0.63 & 0.87 & 0.66 & 0.76 & 0.79 & 0.77 & 0.62 \\
Llama-3.2-3B-Instruct \citep{llama3.2modelcard} & \vline \hspace{0.1cm} 0.78 & \textbf{0.84} & \vline \hspace{0.1cm} 0.62 & \textbf{0.88} & 0.75 & \textbf{0.78 }& 0.81 & 0.86 & 0.62 \\
Gemma-2B \citep{team2024gemma} & \vline \hspace{0.1cm} 0.66 & 0.67 & \vline \hspace{0.1cm} 0.58 & 0.75 & 0.66 & 0.69 & 0.78 & 0.58 & \textbf{0.96} \\
Gemma-2-2B \citep{team2024gemma2} & \vline \hspace{0.1cm} 0.7 & 0.67 & \vline \hspace{0.1cm} 0.64 & 0.8 & 0.74 & 0.7 & 0.75 & 0.63 & 0.95 \\
OLMo-1B-hf \citep{Groeneveld2023OLMo}& \vline \hspace{0.1cm} \textbf{0.83} & \textbf{0.84} & \vline \hspace{0.1cm} \textbf{0.76} & 0.86 & 0.83 & 0.77 & 0.78 & \textbf{0.87} & 0.6 \\
Bloom-1B7 \citep{workshop2022bloom} & \vline \hspace{0.1cm}  0.79 & 0.76 & \vline \hspace{0.1cm} 0.68 & 0.87 & 0\textbf{.85} & 0.77 & 0.62 & 0.71 & 0.84 \\
GPT-Neo-125M \citep{gpt-neo} & \vline \hspace{0.1cm} 0.66 & 0.33 & \vline \hspace{0.1cm} 0.55 & 0.75 & 0.34 & 0.69 & 0.41 & 0.58 & 0.05 \\
GPT2 \citep{radford2019language}  & \vline \hspace{0.1cm} 0.66 & 0.67 & \vline \hspace{0.1cm} 0.57 & 0.76 & 0.53 & 0.72 & 0.77 & 0.6 & 0.93 \\
AlBERT-base-v2 \citep{lan2019albert}& \vline \hspace{0.1cm} 0.7 & 0.68 & \vline \hspace{0.1cm} 0.62 & 0.77 & 0.7 & 0.69 & 0.71 & 0.69 & 0.65 \\
BERT-base-uncased \citep{devlin-etal-2019-bert} & \vline \hspace{0.1cm}  0.79 & 0.83 & \vline \hspace{0.1cm} 0.7 & 0.83 & 0.8 & 0.72 & 0.78 & 0.72 & 0.53 \\

\end{tabular}
}

 \caption{Accuracy for the direction prediction task. Additional terms in the extended stereotype dictionary \citep{Nicolas2021} are embedded and projected to the stereotype dimensions. Projected positive/negative values predict high/low direction, e.g. a value of -0.3 for warmth is registered as low warmth. The highest accuracy for each dimension is shown in bold. Please refer to \autoref{tab:word_counts} for examples of terms with high and low labels for each dimension. } \label{tab:accuracy_table}

\end{table*}

\subsection{Polar Projection}

For the computation of stereotype dimensions, we follow the SensePolar framework by \citet{engler-etal-2022-sensepolar}, which is an extension of the POLAR framework \citep{Mathew2020} for contextual embeddings. Hereby, the embeddings are transformed into an interpretable space based on polar dimensions, which, in our case, are defined by the stereotype content dictionary. We transform the embeddings at two levels: \textbf{(i)} Warmth + competence, and \textbf{(ii)} seven granular dimensions of the extended stereotype content model.

Similar to \citet{fraser-etal-2021-understanding}, we take the words for each stereotype dimension from the theory-driven `seed dictionary' \citep{Nicolas2021}, and we average individually the word embeddings for the high and for the low classified words, for which the numbers are shown in \autoref{tab:word_counts}. Word lists for warmth (sociability + morality) and competence (ability + agency) are compiled of the words of their subordinate dimensions.

As a first step, we calculate the sense embedding $\mathbf{s}$ for a word with its specific sense and \textit{m} sense-specific context examples, as shown in \autoref{equ1}.  We hereby average the contextual embeddings $\mathbf{w}$ across the different context examples, also averaging across subwords when words are split due to subword tokenization. 

\begin{align}\label{equ1}
\mathbf{s} =  \frac{1}{m} \sum_{j=1}^{m} \mathbf{w}_{c_{j}}^{s}   
\end{align}

For \textit{n} words belonging to the pole of a stereotype dimension, e.g. ``friendliness'' and ``sociability'' for the pole ``high sociability'', we average their sense embeddings to obtain the average pole embedding $\mathbf{p}$. Next, we stack the vectors and subtract the low-dimension embeddings from the high-dimension embeddings to obtain the change of basis matrix $\mathbf{a}$, describing the newly defined space with  $\mathbf{h}$ stereotype dimensions:

\begin{align}\label{equ2}
\mathbf{p} = \frac{1}{n} \sum_{i=1}^{n} \mathbf{s}_{i}  \\
\label{equ3} \mathbf{a_h} = \mathbf{p_h}_{high}-\mathbf{p_h}_{low}  
\end{align}

Regarding the warmth and competence transformation, there are only two direction vectors. If, for example, the original contextual embedding has 768 dimensions, the change of basis matrix $\mathbf{a}$ has a shape of (2, 768). 

\vspace{0.1cm}

Before projecting a word of interest to the new dimensions, we compute its embedding $\mathbf{x}$ by again averaging across its context examples as shown in \autoref{equ4}.
Following prior work \citep{engler-etal-2022-sensepolar} we then project the embedding to the new interpretable space by the inverted change of basis matrix as shown in \autoref{eq5}. 

\begin{align}\label{equ4} 
\mathbf{x} = \frac{1}{k} \sum_{i=1}^{k}\mathbf{x}_{c_{i}} \\
 \mathbf{d} = (\mathbf{a}^{T})^{-1}\mathbf{x} \label{eq5} 
\end{align}

The new embedding $\mathbf{d}$ in the 2D or 7D stereotype space can be interpreted as follows: 
Similar to the semantic differential technique, a higher value signifies a higher association with the high pole, for example, ``high morality'', and a lower value signifies a more significant association with the low pole, e.g., ``low morality''. By projecting multiple terms we can compare their differences on these dimensions.

\paragraph{Projection of Additional Terms}

To evaluate the consistency of the stereotype dimensions we follow the approach by \citet{fraser-etal-2021-understanding} and project additional terms from the extended dictionary by \citet{Nicolas2021} to the stereotype space. Hereby, we use the same types of context examples as for the polar space creation. 
For each term, we predict its direction on its assigned dimension by the sign of its polar value, e.g., a value of -0.5 for sociability is registered as low sociability. To calculate the accuracy, we compare these predictions against the labels in the dictionary. 

\paragraph{Projection of Gender-Associated Names \& Gendered Terms}

For the analysis of gender bias, we project gender-associated words to our stereotype dimensions, utilizing two larger binary `vocabulary populations' and individual terms for transgender and nonbinary gender (see \autoref{fig:2d_llama-3}).

The largest populations are 100 historically female-associated names (e.g., Mary, Patricia) and 100 male-associated names (e.g., James, Michael), taken from the most popular given names of the last century in the United States\footnote{https://www.ssa.gov/oact/babynames/decades/century.html}.

Second, we employ binary gendered terms by definition as utilized in experiments of WEAT (Math vs. Arts and Science vs. Art) \citep{Caliskan2017}. For each gender, we project nine terms:

\begin{itemize}
    \item Female terms: female, woman, girl, sister, she, daughter, mother, aunt, grandmother
    \item Male terms: male, man, boy, brother, he, son, father, uncle, grandfather
\end{itemize}

As examples for our gendered terms and names, we use neutral templates, 
such as ``This is [NAME]'' or ``This is [TERM]'' to provide context without unnecessarily introducing additional bias (compare \citet{may-etal-2019-measuring, Tan2019}).
We average across the different templates for a more robust contextual representation of names and terms.

\vspace{0.1cm}
For easier interpretation and comparison between models, we standardize the projected polar values separately for names and terms, as preliminary work showed that named entities and pronouns can show different average tendencies on stereotype dimensions. To assess the significance of the observed differences, we employ t-tests. 

\subsection{Models}

For our evaluation, we project open source models of multiple generations available in the Huggingface Library\footnote{https://huggingface.co/} onto  stereotype dimensions. Model names and references are shown in \autoref{tab:accuracy_table}. Except for the layer-wise analysis, we extract their average contextual representations across all layers, including the first embedding layer.

\section{Results}

\subsection{Prediction of Direction for Additional Terms} \label{prediction}

Predicting the direction on the stereotype dimensions for new terms from the extended stereotype dictionary \citep{Nicolas2021}, we find most studied models can perform this task well, with different strengths.
\autoref{tab:accuracy_table} shows the results when embedding and projecting with the generated examples. 
OlMo-1B-hf (warmth and competence) and Llama3.2-3B-Instruct (only competence) achieve the performance closest to that of static embeddings by \citet{fraser-etal-2021-understanding}, where the FastText-based model scored respectively 0.85 for warmth and 0.86 for competence. 
OlMO and the Llama models also perform very well for the granular dimensions. Predicting morality is the overall easiest task for the models, while the other subdimension of warmth, sociability, poses the most difficult task. Accuracy varies greatly for religion, where there are 142 high, but only 6 low-labeled additional terms.

Only GPT-Neo-125M performs worse than chance for some dimensions, however, this pertains only to raw polar values. When we use a different cut-off than zero for predicting low/high directions by mean-centering the projected values, the model achieves much better results, e.g., 0.76 accuracy for warmth and 0.71 for competence. Similarly, GPT2 and the Gemma models benefit from a mean-based cut-off value, gaining up to 10 percentage points per dimension. Thus projections are spread on different ranges of values, but all models can reasonably discriminate between low and high-labeled terms on the stereotype dimensions.

\begin{figure}
\centering

\includegraphics[width=\linewidth]{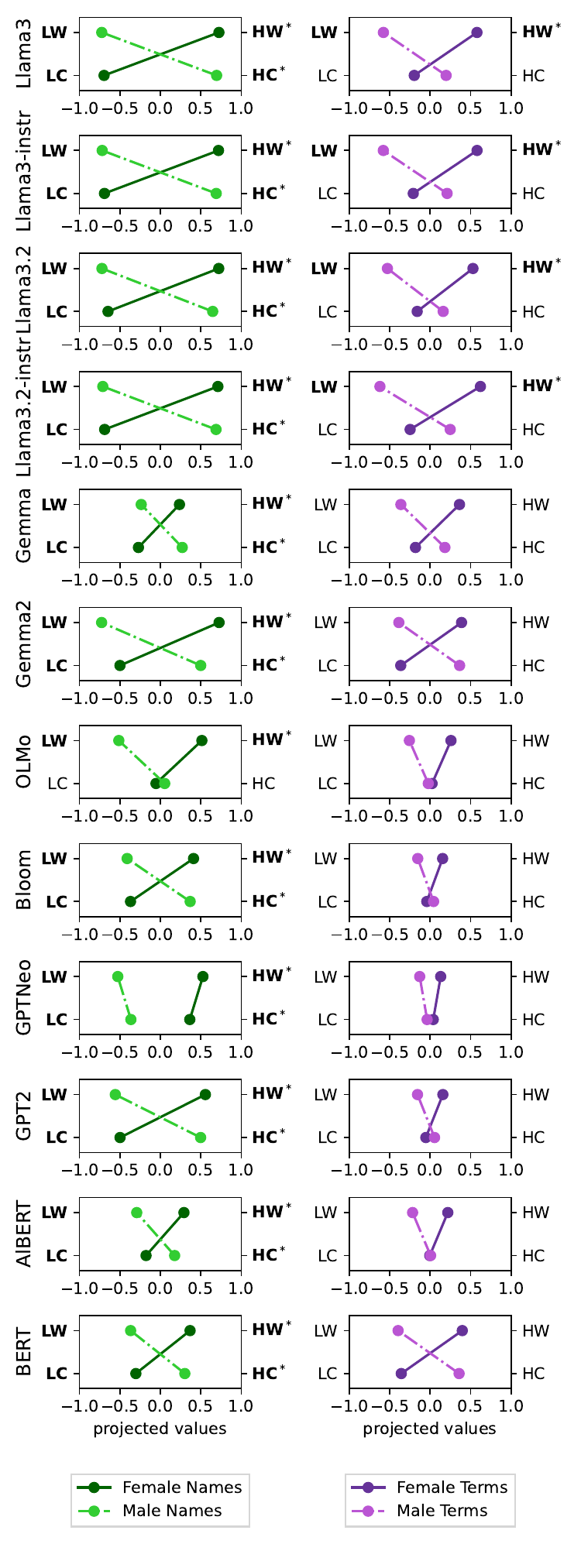}

\caption{2D stereotype profiles for 100 female/male-associated names (left) and 9 female/male gendered terms (right). \\ LW/HW = Low/High Warmth. LC/HC = Low/High Competence. \textbf{*}Dimensions with statistically significant differences (p<0.05).}. \label{fig:2d profiles}
\end{figure}

\subsection{Gender Stereotype Profiles}

For all twelve studied LLMs, we find statistically significant bias for gender-associated names, as evident in the 2D profiles for warmth and competence in \autoref{fig:2d profiles}. In line with human bias found in studies of the stereotype content model \citep{fiske2002model}, the models highly agree on the relative associations of female names with warmth and male names with competence when using the gender non-specific generated contexts. GPTNeo poses an exception, where both dimensions are associated with female names, and for OlMo, the difference in competence is insignificant.

\begin{figure}[h]

\includegraphics[width=0.9\linewidth]{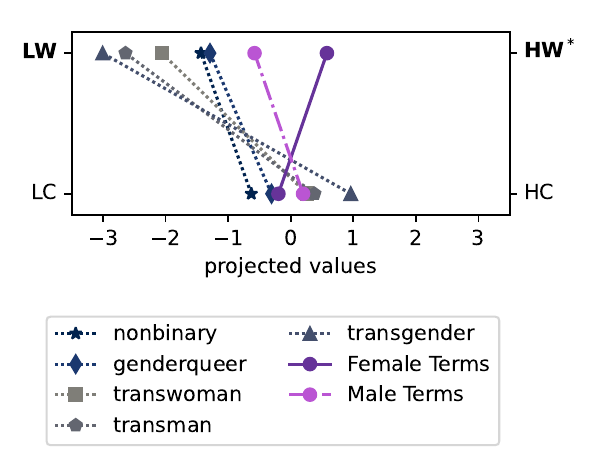}

\caption{2D Stereotype profile for Llama-3-8B (see \autoref{fig:2d profiles}) with additional projections of individual nonbinary terms. LW/HW = Low/High Warmth. LC/HC = Low/High Competence.}. \label{fig:2d_llama-3}
\end{figure}

For the much smaller `vocabulary populations' of gendered terms (nine terms per gender), warmth is the more relevant dimension than competence, with only the former being significantly biased for all four Llama-3 models. For some models, e.g. GPT2, the gendered term differences are small, but the bias direction is very consistent when comparing name and term stereotype profiles. In \autoref{fig:2d_llama-3}, we additionally see the projections of five individual terms beyond binary gender. Nonbinary and transgender-related terms are associated with lower warmth than binary term means, which was found for all newer models (studied variants of LLama3, Gemma, OLMo). For competence, there was no observable trend. For further and statistical analyses of individual terms and small vocabulary populations, future work needs to extend the context examples (as discussed in \autoref{discussion}).

\begin{figure*}[ht]
\centering
\begin{subfigure}{.5\textwidth}
  \centering
  \includegraphics[width=\linewidth]{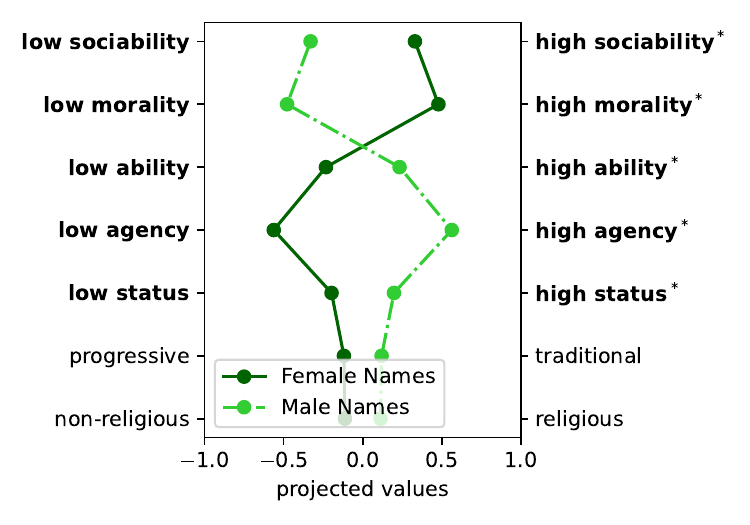}

\end{subfigure}%
\begin{subfigure}{.5\textwidth}
  \centering
  \includegraphics[width=\linewidth]{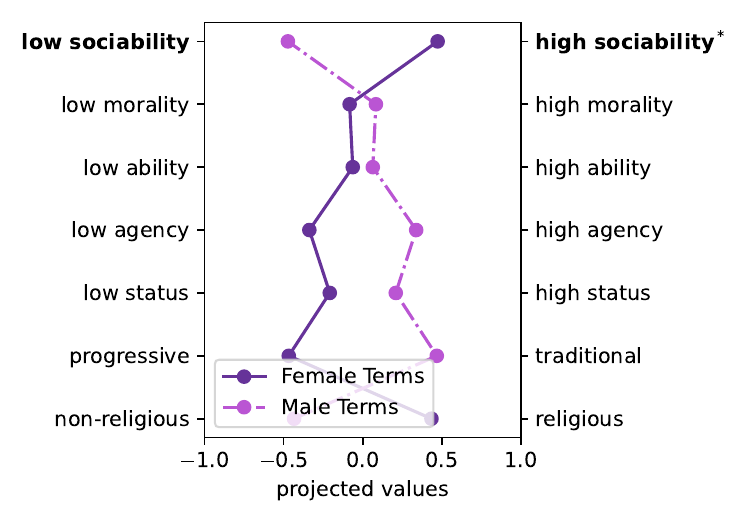}

\end{subfigure}

\caption{
7D stereotype profiles for 100 female/male-associated names (left) and 9 female/male gendered terms (right) for Llama-3.2-3B-instruct.  \textbf{*}Statistically significant differences (p<0.05).}\label{fig:stereotype_profiles_llama3}
\end{figure*}

Zooming into the 7-dimensional stereotype profiles for Lama-3-8B (see \autoref{fig:cover_plot}) and Llama-3.2-3B-instruct in \autoref{fig:stereotype_profiles_llama3}, we can perceive the previously shown gender associations in more detail. The profiles between Llama-3 and the newer and instruction-tuned 3.2 version are quite similar: Sociability and morality (warmth) are linked with female names, which is true for 10 and 11 of all studied models. Ability and agency (competence) are significantly related with male names, which is true for 9 and 7 of the studied models. Also for status, six models show a significant association with male names. Furthermore six models find male names to be more traditional on the political dimension, while there is no perceivable trend for religion.
For the small populations of binary gendered terms, there is only one clearly biased dimension, where seven models agree on a stereotypical association of sociability with female terms.

\begin{figure*}[t]
\centering
\begin{subfigure}{.5\textwidth}
  \centering

    \centering
    \includegraphics[width=\textwidth]{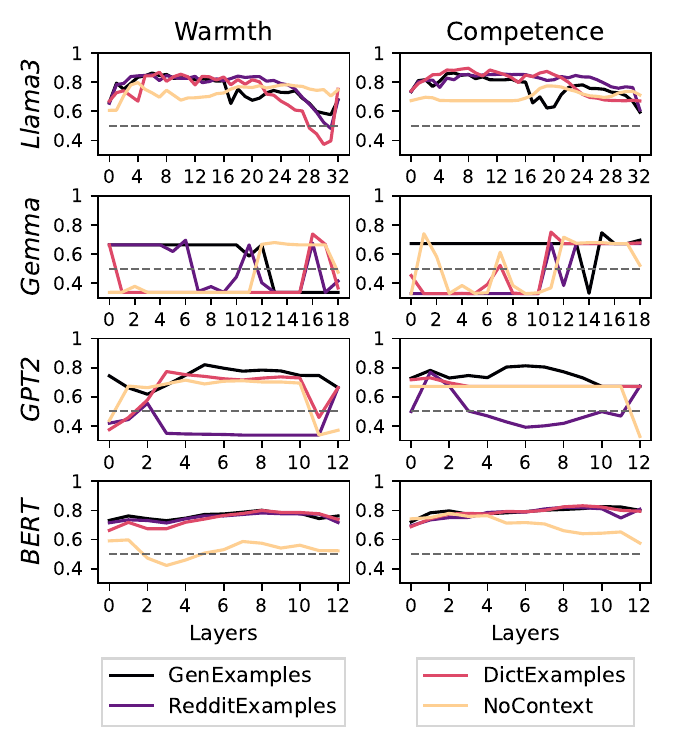}
    \caption{Accuracy for the direction prediction task across layers, with different context examples.     Additional terms are projected to the stereotype dimensions;
    positive/negative values predict high/low direction.}
    \label{fig:warmth_competence_across_layers_main}

\end{subfigure}%
\begin{subfigure}{.5\textwidth}
  \centering
 
    \centering
    \includegraphics[width=\textwidth]{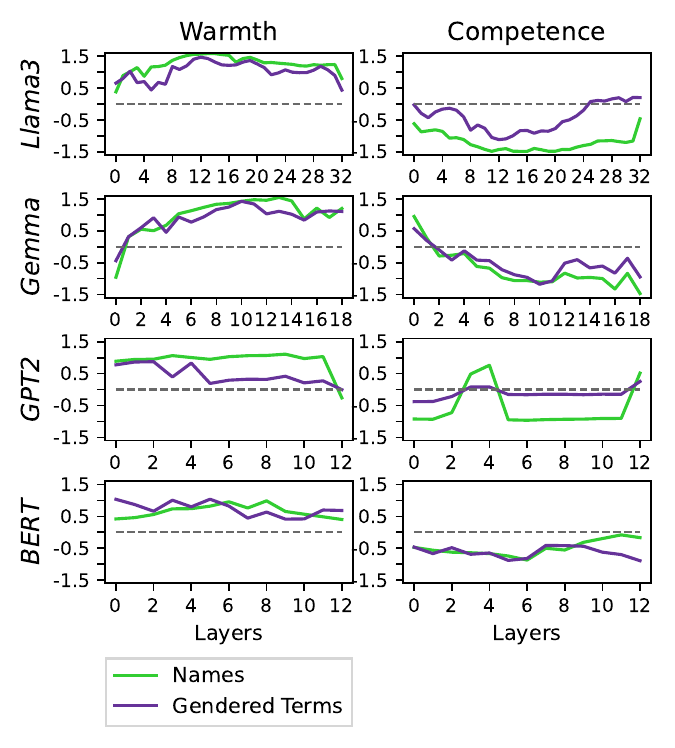}
    \caption{Gender bias across layers with generated gender non-specific examples. Higher values signify bias towards female-associated names/terms; lower values signify bias towards male-associated names/terms.}
    \label{fig:bias_across_layers_main}
  
\end{subfigure}

\caption{Layerwise visualization of prediction accuracy and gender bias for selected models.
} 
\label{fig:across_layers}

\end{figure*}

\subsection{Context Examples and Bias across Layers}

Comparing the performance for generated, dictionary, Reddit examples, and no context across layers in \autoref{fig:warmth_competence_across_layers_main}, we find that the influence of context on the new term prediction task depends on the model. For Llama-3-8B, the accuracy seems quite stable compared to the smaller Gemma-2B variant, where we see high variation across layers and context types. However, as elaborated in \autoref{prediction}, while the cut-off value of zero works well for most, including all Llama models, Gemma could better discriminate between low/high labeled terms by a different cut-off value. For GPT2, the Reddit examples lead to much lower accuracy, while for BERT, the no context condition performs markedly worse. Overall the concepts of warmth and competence behave similarly throughout the layers.

On the right in \autoref{fig:bias_across_layers_main}, we see that bias across layers is rather consistent for all models, where higher values signify bias towards female-associated names/terms and lower values signify bias towards male-associated names/terms. Shown by the example of the generated contexts, stereotypical associations permeate throughout the networks.
In some cases, the first and last layers behave differently, with a reversed bias direction compared to the overall model.

\section{Discussion} \label{discussion}

Our results provide substantial evidence of stereotype dimensions in the embedding space of LLMs and a gender bias that predominantly corresponds to the human bias found in studies of the stereotype content model \citep{fiske2002model}. For all studied models, female names are relatively associated with higher warmth, and for most models, male names are associated with higher competence. 
There is less evidence of bias for the studied gendered terms, which in part is likely due to the small groups of only nine terms per gender. The direction of gender differences is, however, overwhelmingly consistent. We furthermore find stereotype dimensions and bias across layers, in line with prior work that semantics are spread throughout the network \citep{Tenney2019b}.

The projection of contextual embeddings based on the stereotype content model can deliver robust insights when analyzing large vocabulary groups. As the magnitude of the values depends on the properties of the original embedding space, statistical analysis is employed to assert the significance of gender differences. This is viable with the larger collection of gender-associated names, also providing context for the differences between binary gendered terms. For the analysis of small vocabulary groups or individual terms, e.g. for comparing terms of binary and nonbinary gender, increasing the number of context examples offers potential for statistical tests.

While both 2D and 7D stereotype dimensions provide interesting results, a significant gender bias is most evident in the warmth and competence dimensions. These benefit from the larger numbers of low and high-rated words in the dictionaries, increasing the robustness of the concept representations. Likely associations are also more stable when relating to broader concepts.
Therefore, the high-level projection is a suitable first level of analysis and starting point for bias mitigation. 

Significant gender bias, however, may also occur on a more granular level. Different dimensions can be relevant depending on domains and tasks, such as progressive-traditional in the realm of politics, and the mode of projection can be easily adapted with the presented methodology. A combined projection with other dimensions such as valence (unpleasantness vs. pleasantness) (see e.g. \citet{omrani2023evaluating}), could provide even further insights.

As we have shown, the term context can have a considerable influence on the behavior of the stereotype dimensions. Thus, examples for pole and projected terms should be chosen deliberately. Gender non-specific context is our default choice because no additional bias is introduced through the examples and we get a clearer picture of the bias already present within the pre-trained embeddings. Even smaller open-source LLMs are now able to provide examples with this property at scale. However, measurement is certainly best conducted with domain-specific data when a specific use case exists. 
While we use simple templates as contexts for the gendered terms and names, these could as well be sampled from a target domain or be generated to test specific scenarios. For example, similar to \citet{may-etal-2019-measuring}, this could involve introducing success in a historically male-dominated field to the term/name context, to test if a penalty exists for females, as found in psychological studies \citep{Heilman2004}. 

While embedding-based methods for bias measurement have been critiqued for their remoteness from downstream applications \citep{gallegos2024bias}, and are certainly no substitute for task-specific investigations, they have multiple advantages. First, the methodology does not depend on natural language datasets that can be leaked into training data and are therefore applicable to older and newer models alike. Second, the same stereotype dimensions can easily be used for bias mitigation \citep{ungless-etal-2022-robust, omrani-etal-2023-social}, alleviating representational harm and the risk that it influences downstream behavior. Finally, our paper shows they can be exploited for intuitive visualizations exposing gender bias.

\section{Conclusion}

Large pre-trained language models reflect the biases in their training data, which in turn reflect the biases of their creators. As the foundation for AI applications, their biases are further propagated, warranting their study to uncover the risks and promote mitigation efforts. 

In this work, we profile gender stereotypes in twelve LLMs by means of the stereotype content model of social psychology \citep{fiske2002model}, thereby theoretically grounding the analysis, which has in the past been described as the missing link for bias measurements \citep{blodgett-etal-2020-language}. By a matrix transformation, the opaque contextual embeddings of the models reveal interpretable stereotype dimensions. Along the two major dimensions of warmth and competence, we find significant bias for gender-associated names and some evidence of bias for gendered terms, widely aligned with stereotypes found in human studies.

The shown presence of stereotype dimensions in LLMs is a comprehensible replication of semantics in human language, however, the differential associations of social groups along these dimensions constitute a representational harm. 
Equal treatment starts with equal representation; stereotypes already statistically significant in embedding space come with the risk of being exploited in downstream tasks, which could lead to different and unfair treatment of social groups. 
While the first access point would be the training data itself, the embedding space allows a quantification of patterns that is useful for bias assessment and mitigation.
The analysis of embedding spaces by interpretable dimensions provides a means to evaluate both functional semantics and harmful associations that should be `unlearned' to prevent their propagation.

Awareness of bias in LLMs needs to be increased beyond the expert audience, as biased models are already deployed, and completely debiased models may hardly be attainable, as they are trained on vast amounts of biased human-created data. The here presented bias profiles based on the stereotype content model employ an intuitive scoring along meaningful scales of opposing concepts (e.g. low vs. high warmth), as proven effective by semantic differentials in human surveys \citep{Osgood1957}. The result is a highly visual solution for communicating bias to wider audiences and users of artificial intelligence.

\section{Limitations}

The bias profiles presented in this paper concern only gender, but there is a whole range of biases to be profiled in LLMs to evaluate and communicate representational harms. The scope of analysis was also constrained to English contextual embedding spaces and needs to be extended to a multi-lingual setting in the future.

Furthermore, the focus of this paper was binary gender, with historically gender-associated names and gendered terms. While we projected a few terms for transgender and nonbinary gender to the stereotype dimensions, future analysis needs to extend the methodology for these smaller and diverse `vocabulary populations'. Increasing the number of context samples for terms offers potential for greater robustness and applicability of statistical tests. 

Finally, no general bias measurement benchmark or method, including the one presented in this paper, precludes the absolute necessity of task-specific bias measurements. However, they can be a piece of the puzzle by revealing learned general bias tendencies and providing a means to mitigate and communicate these effectively.

\bibliographystyle{acl_natbib}
\bibliography{main}

\end{document}